\documentclass[10pt,twocolumn,letterpaper]{article}

\usepackage{cvpr}              

\usepackage{graphicx}
\usepackage{amsmath}
\usepackage{amssymb}
\usepackage{booktabs}

\usepackage{algorithm}
\usepackage{algorithmic}
\usepackage{color}
\usepackage{siunitx}
\usepackage{bm}
\usepackage{multirow}

\usepackage{amsmath}
\usepackage{amssymb}
\usepackage{booktabs}
\usepackage{mathrsfs}
\usepackage{amssymb}
\usepackage{dsfont}
\usepackage{epstopdf}
\usepackage{subcaption}
\renewcommand{\thefootnote}{\fnsymbol{footnote}} 
\usepackage[switch]{lineno}

\usepackage[pagebackref,breaklinks,colorlinks]{hyperref}

\usepackage[capitalize]{cleveref}
\crefname{section}{Sec.}{Secs.}
\Crefname{section}{Section}{Sections}
\Crefname{table}{Table}{Tables}
\crefname{table}{Tab.}{Tabs.}

\def\cvprPaperID{5060} 
\def\confName{CVPR}
\def\confYear{2023}

\begin{document}
\title{Self-Supervised 3D Scene Flow Estimation Guided by Superpoints}

\author{Yaqi Shen,~Le Hui,~Jin Xie$^{*}$,~Jian Yang\\
	PCA Lab, Nanjing University of Science and Technology, Nanjing, China\\
	\tt\small \{syq, le.hui, csjxie, csjyang\}@njust.edu.cn
}

\maketitle

\let\thefootnote\relax\footnotetext{$^*$Corresponding authors}

\begin{abstract} 
3D scene flow estimation aims to estimate point-wise motions between two consecutive frames of point clouds. Superpoints, i.e., points with similar geometric features, are usually employed to capture similar motions of local regions in 3D scenes for scene flow estimation. However, in existing methods, superpoints are generated with the offline clustering methods, which cannot characterize local regions with similar motions for complex 3D scenes well, leading to inaccurate scene flow estimation. To this end, we propose an iterative end-to-end superpoint based scene flow estimation framework, where the superpoints can be dynamically updated to guide the point-level flow prediction. Specifically, our framework consists of a flow guided superpoint generation module and a superpoint guided flow refinement module. In our superpoint generation module, we utilize the bidirectional flow information at the previous iteration to obtain the matching points of points and superpoint centers for soft point-to-superpoint association construction, in which the superpoints are generated for pairwise point clouds. With the generated superpoints, we first reconstruct the flow for each point by adaptively aggregating the superpoint-level flow, and then encode the consistency between the reconstructed flow of pairwise point clouds. Finally, we feed the consistency encoding along with the reconstructed flow into GRU to refine point-level flow. Extensive experiments on several different datasets show that our method can achieve promising performance. Code is available at \emph{\url{https://github.com/supersyq/SPFlowNet}}.
\end{abstract}

\section{Introduction}
Scene flow estimation is one of the vital components of numerous applications such as 3D reconstruction \cite{gotardo2015photogeometric}, autonomous driving \cite{luo2021self}, and motion segmentation \cite{baur2021slim}. 
Estimating scene flow from stereo videos and RGB-D images has been studied for many years \cite{huguet2007variational,hornacek2014sphereflow}. Recently, with the rapid development of 3D sensors, estimating scene flow from two consecutive point clouds has receiving more and more attention. However, due to the irregularity and sparsity of point clouds, scene flow estimation is still a challenging problem in real scenes.

\begin{figure}[t]
	\centering
	\includegraphics[width=0.48\textwidth]{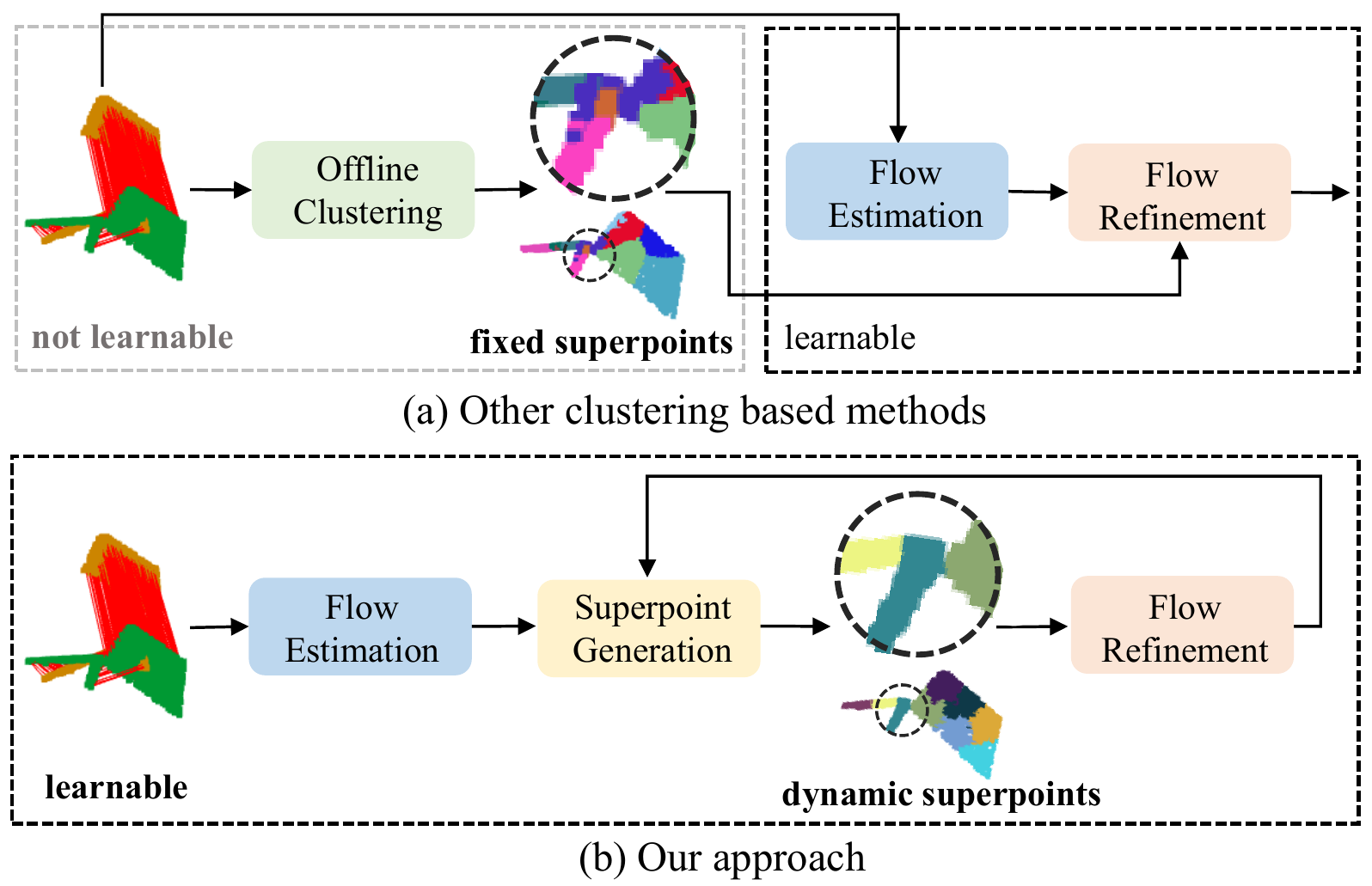}
	\vspace{-15pt}
	\caption{Comparison with other clustering-based methods. (a) Other clustering based methods utilize offline clustering algorithms to split the point clouds into some fixed superpoints for subsequent flow refinement, which is not learnable. (b) Our method embeds the differentiable clustering (superpoint generation) into our pipeline and generates dynamic superpoints at each iteration. We visualize part of the scene in FlyingThings3D \cite{mayer2016large} for better visualization. Different colors indicate different superpoints and red lines indicate the ground truth flow.}
	\vspace{-5pt}
	\label{image1}
\end{figure}

In recent years, many 3D scene flow estimation methods have been proposed \cite{puy2020flot,luo2021self,gu2022rcp,wang2022estimation,wei2021pv,shi2022safit,jiang20223d}. Most of these methods \cite{liu2019flownet3d, puy2020flot, wang2022estimation} regress the flow via the point-level correlations between the point clouds. Although these methods achieve promising results, they ignore high-level cues, i.e., similar flow patterns at the region level, thereby generating inconsistent flow for the local regions. 
Therefore, some efforts have been dedicated to explicitly encoding the consistent flow constraints at the superpoint level for scene flow estimation \cite{gojcic2021weakly,li2021hcrf,li2022rigidflow,dong2022exploiting}. 
For example, HCRF-Flow \cite{li2021hcrf} uses offline oversegmentation method \cite{lin2018toward} to split the scenes into superpoints and takes the superpoints as rigid objects for flow refinement with conditional random fields.
Besides, Rigid3DSceneFlow \cite{gojcic2021weakly} and LiDARSceneFlow \cite{dong2022exploiting} utilize the DBSCAN clustering algorithm \cite{ester1996density} to segment the estimated foreground points into superpoints for flow rigidity constraints.
In addition, RigidFlow \cite{li2022rigidflow} utilizes \cite{lin2018toward} to decompose the point clouds into fixed superpoints and estimates the rigid transformations for superpoints as pseudo scene flow labels for model training. 
In summary, these clustering based methods utilize offline clustering algorithms with hand-crafted features (i.e., coordinates and normals) to generate the superpoints and use the consistent flow constraints on these fixed superpoints for scene flow estimation. However, for some complex scenes, the offline clustering methods may cluster points with different flow patterns into the same superpoints. Figure \ref{image1}(a) shows that \cite{lin2018toward} falsely clusters points with the entirely different flow into the same superpoint colored in purple (highlighted by the dotted circle). 
Thus, applying flow constraints to the incorrect and fixed superpoints for flow estimation will mislead the model to generate false flow results.

To address this issue, we propose an iterative end-to-end superpoint guided scene flow estimation framework (dubbed as ``SPFlowNet''), which consists of an online superpoint generation module and a flow refinement module. 
Our pipeline jointly optimizes the flow guided superpoint generation
and superpoint guided flow refinement for more accurate flow prediction
(Figure \ref{image1}(b)).
Specifically, we first utilize farthest point sampling (FPS) to obtain the initial superpoint centers, including the coordinate, flow, and feature information. 
Then, we use the superpoint-level and point-level flow information in the previous iteration to obtain the matching points of points and superpoint centers. With the pairs of points and superpoint centers, we can learn the soft point-to-superpoint association map. And we utilize the association map to adaptively aggregate the coordinates, features, and flow values of points for superpoint center updating. 
Next, based on the updated superpoint-wise flow values, we reconstruct the flow of each point via the generated association map. Furthermore, we encode the consistency between the reconstructed flow of pairwise point clouds. 
Finally, we feed the reconstructed flow along with the consistency encoding into a gated recurrent unit to refine the point-level flow. 
Extensive experiments on several benchmarks show that our approach achieves state-of-the-art performance. 

Our main contributions are summarized as follows:
\begin{itemize}
	\item We propose a novel end-to-end self-supervised scene flow estimation framework, which iteratively generates dynamic superpoints with similar flow patterns and refines the point-level flow with the superpoints.
	\item Different from other offline clustering based methods,
	we embed the online clustering into our model to dynamically segment point clouds with the guidance from pseudo flow labels generated at the last iteration.
	\item A superpoint guided flow refinement layer is introduced to refine the point-wise flow with superpoint-level flow information, where the superpoint-wise flow patterns are adaptively aggregated into the point-level with the learned association map.
	\item Our self-supervised scene flow estimation method outperforms state-of-the-art methods by a large margin.
\end{itemize}

\section{Related Work}
\noindent \textbf{Supervised scene flow estimation on point clouds.}
The scene flow describes the 3D displacements of points between two temporal frames \cite{vedula1999three}. Estimating scene flow from stereo videos and RGB-D images has been investigated for many years \cite{huguet2007variational,herbst2013rgb,jaimez2015primal,teed2021raft}. Recently, with the development of 3D sensor, directly estimating scene flow on point clouds has drawn the interest of many researchers. 
There are some supervised scene flow estimation methods \cite{behl2019pointflownet,wang2020flownet3d++,wei2021pv,wang2021hierarchical,jund2021scalable,wang2022ThreeDFlow}.
FlowNet3D \cite{liu2019flownet3d} is the first end-to-end scene flow estimation framework on point clouds with a flow embedding layer to capture the local correlation between source and target point clouds and a set upconv layer to propagate the flow embedding from the coarse scale to the finer scale for flow regression. 
Except for FlowNet3D, some other methods also involve multiscale analysis, such as \cite{gu2019hplflownet,wang2021festa,wang2021hierarchical, wang2022residual,cheng2022Bi}. Among them, PointPWC-Net builds a coarse-to-fine scene flow estimation framework with novel cost volume, upsampling and warping layers, and the cost volume layer only involves local neighborhoods.
Besides, Bi-PointFlowNet \cite{cheng2022Bi} propagates the features of two frames bidirectionally at different scales to obtain bidirectional correlations, which achieves promising performance. 
Nevertheless, the above methods build local correlations within a limited search area, which fail to accurately estimate the large displacements. Therefore, FLOT \cite{puy2020flot} and SCTN \cite{li2021sctn} adopt optimal transport to build global correlation. 
In contrast, CamLiFlow \cite{liu2022camliflow} takes two consecutive synchronized camera and Lidar frames as inputs to estimate the optical flow and scene flow simultaneously and builds multiple bidirectional connections between its 2D and 3D branches to fuse the information of two modalities. 
Unlike other methods that focus on a pair of point clouds, SPCM-Net \cite{he2022learning}, MeteorNet \cite{liu2019meteornet}, and \cite{huang2022accumulation} take a sequence of point clouds as input. Specifically, SPCM-Net computes spatiotemporal cost volumes between pairwise two frames and utilizes an order-invariant recurrent unit to aggregate the correlations across time.
Although these supervised scene flow estimation methods achieve adorable performance, they need dense supervision for model training, while acquiring point-wise annotations is expensive.

\noindent \textbf{Self-supervised scene flow estimation on point clouds.}
To address this drawback, there are some self-supervised and weakly-supervised methods \cite{zuanazzi2020adversarial,ouyang2021occlusion2,lu20213d,luo2021self,li2022learning, he2022self}. 
The self-supervised methods \cite{mittal2020just, pontes2020scene,tishchenko2020self} utilize the cycle-consistency loss and nearest neighbor loss for model training.
Besides, PointPWC-Net \cite{wu2020pointpwc} combines the nearest neighbor loss with a flow smoothness loss and a Laplacian regularization loss as the self-supervised loss. \cite{li2021self} generates pseudo labels by optimal transport and refines the generated pseudo labels with the random walk. The generated pseudo labels are used for unsupervised model optimization. The follow-up RigidFlow \cite{li2022rigidflow} utilizes optimization-based point cloud oversegmentation method \cite{lin2018toward} to split point clouds into a set of supervoxels and then calculates the rigid transformation as pseudo flow labels. Rigid3DSceneFlow \cite{gojcic2021weakly} and LiDARSceneFlow \cite{dong2022exploiting} get rid of the requirement for expensive point-wise flow supervision with binary background masks as well as ego-motion and utilize the DBSCAN clustering algorithm \cite{ester1996density} to segment the foreground points for flow rigidity constraints.
LiDARSceneFlow expands \cite{gojcic2021weakly} with a Gated Recurrent Unit (GRU) for flow refinement. 
The previous methods based on offline clustering mainly employ hand-crafted features (i.e., coordinates and normals) to offline cluster superpoints, which may cluster points with different motion patterns into the same clusters and further lead to worse results with rigidity constraints on the incorrect clusters. Our method attempts to dynamically cluster point clouds into superpoints and then refines the point-wise flow with superpoint-level flow information. In this way, our model can jointly optimize the superpoint generation and flow refinement for more accurate results. Additionally, other self-supervised methods \cite{gu2022rcp,jin2022deformation,battrawy2022rms} also achieve promising performance.

\noindent \textbf{Point cloud oversegmentation.} Point cloud oversegmentation semantically clusters points into superpoints. 
Recently, some optimization-based superpoint oversegmentation methods are proposed \cite{guinard2017weakly,lin2018toward}. Among them, \cite{lin2018toward} converts the point cloud oversegmentation into a subset selection problem and develops a heuristic algorithm to solve it.  
In contrast, SPNet \cite{hui2021superpoint} is the first end-to-end superpoint generation network. Due to low computational cost, superpoints are used for many down-stream tasks, such as point cloud segmentation \cite{cheng2020cascaded,cheng2021sspc,Tang_2022_ACCV,hui2022learning}. In this paper, we introduce superpoints into scene flow estimation based on SPNet. Different from that SPNet focuses on generating superpoints in a single point cloud, our model utilizes the bidirectional flow information at the previous iteration to guide superpoint generation for pairwise point clouds.

\begin{figure*}[thb]
	\begin{center}
		\includegraphics[width=\linewidth]{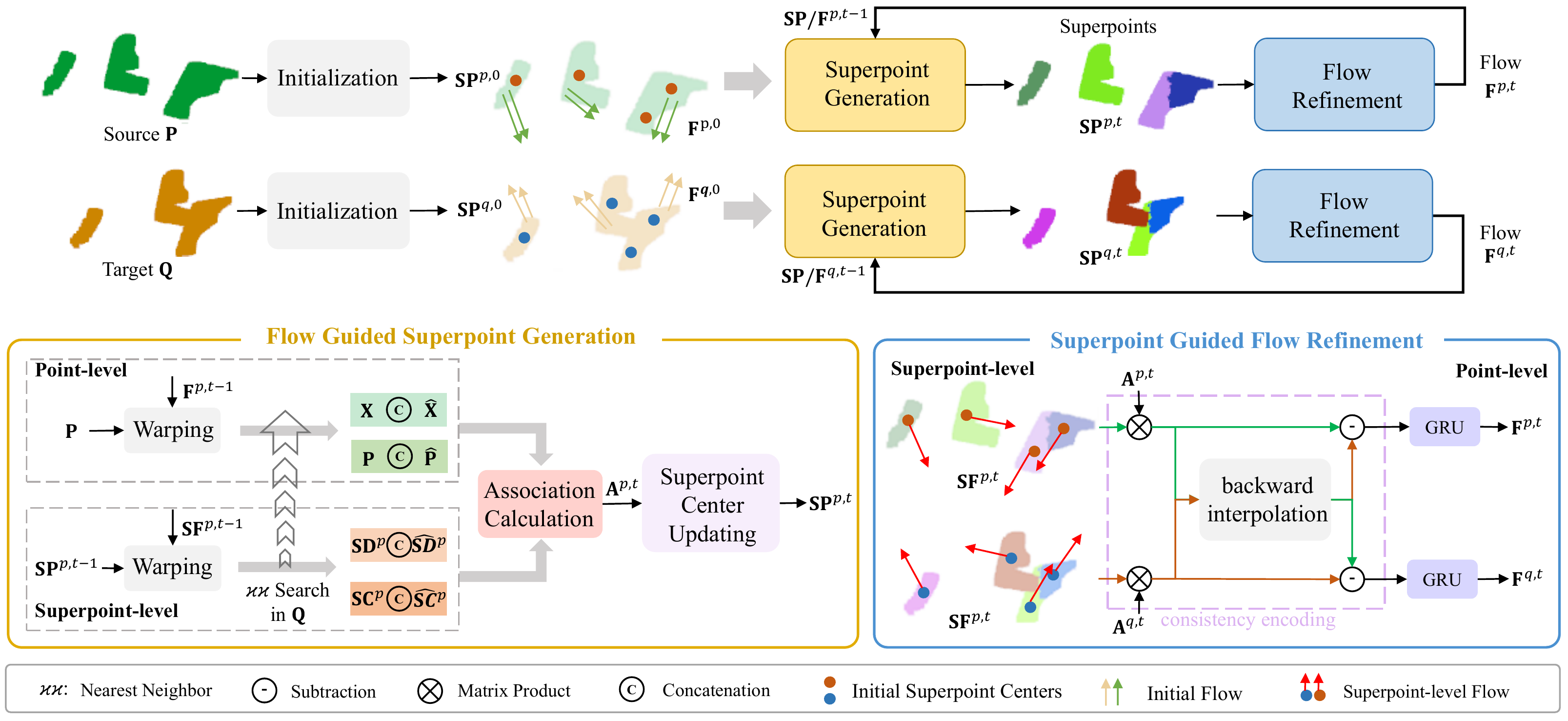}
	\end{center}
	\caption{An overview of SPFlowNet. Given two consecutive point clouds  $\mathbf{P}$ and $\mathbf{Q}$, we first calculate the initial flow $\mathbf{F}^0 = \{\mathbf{F}^{p,0}, \mathbf{F}^{q,0}\}$ and the initial superpoint centers $\mathbf{SP}^0 = \{\mathbf{SP}^{p,0}, \mathbf{SP}^{q,0}\}$ at the initialization stage ($t = 0$). Then, our model iteratively performs the flow guided superpoint generation module and the superpoint guided flow refinement module for scene flow estimation. In the end, we can obtain final flow results after several iterations.
	Specifically, at the $t$-th iteration, the flow guided superpoint generation module clusters points into dynamic superpoints $\mathbf{SP}^t = \{\mathbf{SP}^{p,t}, \mathbf{SP}^{q,t}\}$ with the pseudo superpoint-level and point-level flow labels generated at the previous iteration. With the generated superpoints, the superpoint guided flow refinement module feeds the superpoint-level flow and consistency encoding into GRU to obtain the updated point-level flow $\mathbf{F}^{p,t}$ and $\mathbf{F}^{q,t}$. 
	}
	\label{model}
\end{figure*}

\section{Method}
In this section, we illustrate our superpoint guided scene flow estimation (SPFlowNet) framework in detail. As shown in Figure \ref{model}, 
SPFlowNet consists of a flow guided superpoint generation module and a superpoint guided flow refinement module.
It takes two consecutive point clouds $\mathbf{P} = \{\mathbf{p}_i\in\mathbb{R}^3\mid i= 1,2,\ldots,n\}$ and $\mathbf{Q} = \{\mathbf{q}_j \in\mathbb{R}^3 \mid j=1,2,\ldots,m\}$ as inputs and outputs the flow $\mathbf{F}^t = \{\mathbf{F}^{p,t}, \mathbf{F}^{q,t}\}$  at the $t$-th iteration for point clouds $\mathbf{P}$ and $\mathbf{Q}$, respectively. 
Note that the iteration subscript $t = 0$ means that our model is in the initialization stage.

\subsection{Initialization}\label{initial}
\noindent\textbf{Initial flow.} 
Firstly, we utilize the feature encoder used in FLOT \cite{puy2020flot} to extract the features for point clouds $\mathbf{P}$ and $\mathbf{Q}$. The local features of $\mathbf{P}$ and $\mathbf{Q}$ can be denoted as $ \mathbf{X} \in \mathbb{R}^{n \times d} $ and $\mathbf{Y} \in \mathbb{R}^{m \times d}$, where $d$ is the dimension of the feature. 
Then, we calculate the global correlation $\mathbf{W} \in \mathbb{R}^{n \times m}$ between the point clouds $\mathbf{P}$ and $\mathbf{Q}$, where $\mathbf{W}$ can be formulated as the dot product between their features. Next, we apply the Sinkhorn algorithm \cite{sinkhorn1964relationship} to it for the final correlation map $\mathbf{W}$.  
The initial flow $ \mathbf{f}_{i}^{p,0} \in \mathbf{F}^{p,0}$ for each point $ \mathbf{p}_{i} \in \mathbf{P}$ can be defined as
\begin{equation}\label{0}
\mathbf{f}_{i}^{p,0} = \frac{\sum\nolimits_{j=1}^{m}w_{i,j}\mathbf{q}_j}{\sum\nolimits_{j=1}^{m}w_{i,j}} - \mathbf{p}_i
\end{equation}
Similarly, we can obtain the initial flow $\mathbf{F}^{q,0}$ for point clouds $\mathbf{Q}$ by taking the same operations as $\mathbf{P}$ on $\mathbf{Q}$.

\noindent\textbf{Initial superpoint center.} We obtain $L$  ($L \ll n $ and $L \ll m$) initial superpoint centers $\mathbf{SP}^0 = \{\mathbf{SP}^{p,0}, \mathbf{SP}^{q,0}\}$ for point clouds $\mathbf{P}$ and $\mathbf{Q}$ by employing the FPS algorithm in the coordinate space. $\mathbf{SP}^{p,0}$ and $\mathbf{SP}^{q,0}$ denote the initial superpoint centers for pairwise point clouds $\mathbf{P}$ and $\mathbf{Q}$, respectively. Each superpoint center includes the coordinate, flow, and descriptor information, denoted by $\mathbf{SC}^0$, $\mathbf{SF}^0$, and $\mathbf{SD}^0$, respectively. 

\subsection{Flow Guided Superpoint Generation}\label{superpoint}
The scene flow estimation methods \cite{gojcic2021weakly,li2022rigidflow} usually exploit the offline clustering methods \cite{ester1996density,lin2018toward} to decompose the point clouds into a collection of clusters and employ the flow rigidity constraints on the fixed clusters. However, the offline clustering methods usually generate false clusters, where the points with different flow patterns exist in the same cluster, as shown in Figure \ref{image1}(a). Therefore, an online flow guided superpoint generation module is embedded in our framework, in which the point clouds are dynamically divided into superpoints. Due to the joint end-to-end optimization with the consequent flow refinement module, our model can relieve the above problem to some extent.

\noindent\textbf{Point-to-Superpoint association calculation.}
Our method attempts to generate superpoints that satisfy the following requirements: (1) The points of the same superpoint are with similar flow patterns; (2) They are also close to the superpoint centers in the coordinate space; (3) Their features are semantically similar with each other.
Thus, we follow SPNet \cite{hui2021superpoint} to build the soft association map between points and superpoint centers by adaptively learning the bilateral weights from both the coordinate and feature spaces. 
Different from SPNet, we introduce the previously iterated flow information at both point level and superpoint level to obtain the corresponding point/superpoint center via the bidirectional warping operation (source $\rightarrow$ target and target $\rightarrow$ source). Thus, we employ pairs of points and superpoint centers in the source and target point clouds to learn the similarity across the source and target while SPNet does not consider the pairs of corresponding points to learn the similarity. Note that following SPNet, we only calculate the association weights between each point and its $K$-nearest superpoint centers ($K \ll L$) in the coordinate space, which is more efficient.

We take the source point cloud $\mathbf{P}$ as an example to illustrate the point-to-superpoint association map calculation. Specifically, for the $i$-th point in source point cloud $\mathbf{P}$, we use the Euclidean distance in the coordinate space to select the attended $K$ superpoint centers $\mathcal{N}_i$, where
$\mathcal{N}_i = \{ \mathbf{sc}^{p,t-1}_{i,k}\in\mathbb{R}^3,  \mathbf{sf}^{p,t-1}_{i,k}\in\mathbb{R}^3 , \mathbf{sd}^{p,t-1}_{i,k}\in\mathbb{R}^d \}_{k=0}^K$ includes the coordinate, flow, and feature information of the $K$-nearest superpoint centers.
At the $t$-th iteration, the association ${a}^{p,t}_{i,k} \in \mathbf{A}^{p,t}$ between the $i$-th point and the $k$-th superpoint center in source point cloud $\mathbf{P}$ is defined as
\begin{equation}\label{1}
\begin{split}
{a}^{p,t}_{i,k} &= \operatorname{MLP} \left( \mathbf{u}_{i,k} \right) + \operatorname{MLP} \left(\mathbf{g}_{i,k} \right)\\
\mathbf{u}_{i,k} &= \left(\mathbf{x}_i || \mathbf{\hat x}^{t-1}_i\right) -\left(\mathbf{sd}^{p,t-1}_{i,k} || \mathbf{\hat {sd}}^{p,t-1}_{i,k} \right)\\
\mathbf{g}_{i,k} &= \left(\mathbf{p}_i || \mathbf{\hat p}^{t-1}_i \right) -\left(\mathbf{sc}^{p,t-1}_{i,k} || \mathbf{\hat {sc}}^{p,t-1}_{i,k} \right)
\end{split}
\end{equation}
where $||$ is the concatenation, 
$\mathbf{u}_{i,k} \in \mathbb{R}^{1 \times(2*d)}$ and $\mathbf{g}_{i,k} \in \mathbb{R}^{1  \times (2*3)}$ represent the differences between the $i$-th point and the $k$-th superpoint center in feature and coordinate spaces, respectively. Besides, $\operatorname{MLP}(\cdot)$ denotes a multi-layer perceptron followed by a sum-pooling operation, which is used to map the above difference information to association weights in both coordinate and feature spaces.

In Equation (\ref{1}), we also utilize the feature and coordinate information of their corresponding points generated by the predicted point-level and superpoint-level flow in the previous iteration. 
The corresponding point $(\mathbf{\hat p}^{t-1}_i, \mathbf{\hat x}^{t-1}_i)$ and superpoint center $(\mathbf{\hat {sc}}^{p,t-1}_{i,k}, \mathbf{\hat {sd}}^{p,t-1}_{i,k})$ for point $\mathbf{p}_i$ and superpoint center $\mathbf{sc}^{p,t-1}_{i,k}$ are defined as
\begin{equation}\label{2}
\begin{split}
\mathbf{\hat p}^{t-1}_i &= \mathbf{p}_i + \mathbf{f}_i^{p,t-1},\mathbf{\hat {sc}}^{p,t-1}_{i,k} = \mathbf{sc}^{p,t-1}_{i,k} + \mathbf{sf}^{p,t-1}_{i,k}\\
 \mathbf{\hat x}^{t-1}_i &= \mathbf{Y}_{\operatorname{NN}(\mathbf{\hat p}^{t-1}_i, \mathbf{Q})}, \mathbf{\hat {sd}}^{p,t-1}_{i,k}= \mathbf{Y}_{\operatorname{NN}(\mathbf{\hat{sc}}^{p,t-1}_{i,k}, \mathbf{Q})}
\end{split}
\end{equation}
where $\operatorname{NN}(\cdot,\mathbf{Q})$ is used to obtain the index of the nearest matching point in target point cloud $\mathbf{Q}$.

Next, we assign each point $\mathbf{p}_{i} \in \mathbf{P}$ a probability vector over its $K$-nearest superpoint centers by 
\begin{equation}
a^{p,t}_{i,k} = \operatorname{softmax} \left(\left[{a}^{p,t}_{i,1},...,{a}^{p,t}_{i,K})\right]\right)_k
\end{equation}
Similarly, we can obtain the association map $\mathbf{A}^{q,t}$ between the target point cloud $\mathbf{Q}$ and its superpoint centers.
 
\noindent\textbf{Superpoint center updating.}
With the generated association map $\mathbf{A}^t$, we can assign each point to its $K$-nearest superpoint centers with the learned weights. For each superpoint center, we adaptively aggregate the coordinate, flow, and feature information of the points belonging to it to update this superpoint center via the normalized association map. Specifically, given the local feature $\mathbf{X}$, flow $\mathbf{F}^{p,t-1} = \{\mathbf{f}^{p,t-1}_i|i=1,\dots,n\}$ at the iteration $t-1$ and the association map $\mathbf{A}^{p,t}$ at the current $t$-th iteration of the source point cloud $\mathbf{P}$, the updated $l$-th superpoint center in source point clouds can be formulated as 
\begin{equation}
\begin{split}
\mathbf{sc}^{p,t}_{l} &= \frac{1}{r}\sum\nolimits_{i=1}^n \mathds{1}\left\{ l \in \mathcal{N}_i \right\}a^{p,t}_{i,l}\mathbf{p}_i \\
\mathbf{sf}^{p,t}_{l} &= \frac{1}{r}\sum\nolimits_{i=1}^n \mathds{1}\left\{ l \in \mathcal{N}_i \right\}a^{p,t}_{i,l}\mathbf{f}^{p,t-1}_i \\
\mathbf{sd}^{p,t}_{l} &= \frac{1}{r}\sum\nolimits_{i=1}^n \mathds{1}\left\{ l \in \mathcal{N}_i \right\}a^{p,t}_{i,l}\mathbf{x}_i 
\end{split}
\end{equation}
where $\mathds{1}\left\{ l \in \mathcal{N}_i \right\}$ is an indicator function that equals to one if the $l$-th superpoint center belongs to $\mathcal{N}_i$, and zero otherwise. Besides, $r=\sum\nolimits_{i=1}^n \mathds{1}\left\{ l \in \mathcal{N}_i \right\}a^{p,t}_{i,l}$ is the normalization factor. Similarly, we update the superpoint centers in target point cloud $\mathbf{Q}$. 
For brevity, we only visualize the pipeline of flow guided superpoint generation for source point cloud $\mathbf{P}$ in Figure \ref{model}. 

\subsection{Superpoint Guided Flow Refinement}\label{refinement}
Inspired by RAFT \cite{teed2020raft}, many scene flow estimation methods \cite{wei2021pv,kittenplon2021flowstep3d,gu2022rcp} utilize a Gate Recurrent Unit (GRU) to iteratively update the predicted flow.

\noindent\textbf{Gated recurrent unit.} Given the hidden state $\mathbf{h}^{t-1}$ at the iteration $t-1$ and the current iteration information $\mathbf{v}^t$,
the calculations of GRU can be written as
\begin{equation}\label{5}
\begin{split}
\mathbf{z}^t&=\sigma \left(\operatorname{SetConv}_z \left(\mathbf{h}^{t-1}||\mathbf{v}^t \right) \right) \\
\mathbf{r}^t&=\sigma \left(\operatorname{SetConv}_r\left(\mathbf{h}^{t-1}||\mathbf{v}^t\right) \right) \\
\mathbf{\hat h}^t&=\operatorname{tanh} \left(\operatorname{SetConv}_h\left(\left(\mathbf{r}^{t} \odot \mathbf{h}^{t-1} \right)||\mathbf{v}^t\right) \right) \\
\mathbf{h}^t&=\left( 1 - \mathbf{z}^t \right) \odot\mathbf{h}^{t-1} + \mathbf{z}^t\odot\mathbf{\hat h}^{t}
\end{split}
\end{equation}
where $\odot$ is the Hadamard product, $||$ is the concatenation, and $\sigma(\cdot)$ is the sigmoid function. The $\operatorname{SetConv}$ layers are adopted from \cite{kittenplon2021flowstep3d,puy2020flot}. 

The existing GRU-based methods usually concatenate the feature, flow, and flow embedding of each point as current iteration information $\mathbf{v}^{t}$, and regress the flow from the new hidden state $\mathbf{h}^{t}$. Although these methods achieve promising results, most of them only involve point-level flow information. In contrast, \cite{dong2022exploiting} converts GRU output into rigid flow according to pre-clustered local regions, it is limited by pre-clustered regions. Our method adaptively learns the flow association at the superpoint level and does not rely on rigid object assumption. Specifically, we encode the superpoint-level flow information into the current iteration information $\mathbf{v}^t$ to guide the new hidden state $\mathbf{h}^t$ generation. Moreover, we utilize the consistency between the reconstructed flow values from the generated superpoints of pairwise point clouds to encode the confidence into the current iteration information $\mathbf{v}^t$. 
Therefore, the current iteration information in our model simultaneously considers the superpoint-level flow information and its confidence.

\noindent\textbf{Superpoint-level flow reconstruction.} 
With the updated superpoint-level flow $\mathbf{SF}^{p,t}$ and $\mathbf{SF}^{q,t}$ for superpoint centers in both point clouds $\mathbf{P}$ and $\mathbf{Q}$, here we map the superpoint-level flow of $K$-nearest superpoint centers back onto each point in original point clouds via the learned association map $\mathbf{A}^{t}$ as follows
\begin{equation}
\mathbf{\widetilde F}^{p,t}_{i} = \sum\nolimits_{k=1}^K {a}^{p,t}_{i,k}\mathbf{sf}^{p,t}_k, \mathbf{\widetilde F}^{q,t}_{i} = \sum\nolimits_{k=1}^K {a}^{q,t}_{i,k}\mathbf{sf}^{q,t}_k
\end{equation}
where $\mathbf{\widetilde F}^{p,t}_{i}$ and $\mathbf{\widetilde F}^{q,t}_{i}$ are the reconstructed superpoint-level flow for point clouds $\mathbf{P}$ and $\mathbf{Q}$, respectively.
In this way, the reconstructed flow of each point in original point clouds adaptively aggregates the superpoint-level flow values of its $K$-nearest superpoint centers. Since the superpoint-level flow values capture the flow patterns of the generated superpoints, we aim to utilize the superpoint-level flow pattern to guide the point-level flow refinement.
\noindent\textbf{Consistency encoding.} 
We do a backward interpolation $\Omega$ used in \cite{ouyang2021occlusion} to propagate the reconstructed superpoint-level flow in the source point clouds to the target point clouds and vice versa. Next, we utilize the consistency between the interpolated flow and reconstructed flow to encode the confidence of the superpoint-level flow by
\begin{equation}
\mathbf{C}^{p,t} =  \pi\left(\mathbf{\widetilde F}^{p,t} - \Omega(\mathbf{\widetilde F}^{q,t} )\right),
\mathbf{C}^{q,t} = \tau\left(\mathbf{\widetilde F}^{q,t} - \Omega(\mathbf{\widetilde F}^{p,t})\right)
\end{equation}
where $\pi$ and $\tau$ are the MLP layers with a sigmoid function.

Besides, we send the reconstructed superpoint-level flow $\mathbf{\widetilde F}^{p,t}$ into a flow embedding layer used in \cite{kittenplon2021flowstep3d} to obtain the correlation feature $\mathbf{F}^{p,t}_c$ and a Linear layer to encode the flow feature $\mathbf{F}^{p,t}_e$.    
With the confidence $\mathbf{C}^{p,t}$, the current iteration information $\mathbf{v}^t$ for source point cloud $\mathbf{P}$ can be defined as
\begin{equation}
\mathbf{v}^t =  \operatorname{SetConv_c}\left(\mathbf{F}^{p,t}_c\mathbf{C}^{p,t}  \right) + \operatorname{SetConv_e}\left(\mathbf{F}^{p,t}_e\mathbf{C}^{p,t}  \right) 
\end{equation}
where the $\operatorname{SetConv}$ layers are adopted from \cite{kittenplon2021flowstep3d,puy2020flot}.

We send $\mathbf{v}^t$ into GRU to obtain the new hidden state $\mathbf{h}^t$. Finally, given the new hidden state $\mathbf{h}^t$, we use a flow regressor to obtain the residual flow $\triangle\mathbf{F}^{p,t}$. Therefore, the updated flow for source point cloud $\mathbf{P}$ at the iteration $t$ can be formulated as $\mathbf{F}^{p,t} = \mathbf{\widetilde F}^{p,t-1} + \triangle\mathbf{F}^{p,t}$. Similarly, we can obtain the updated flow $\mathbf{F}^{q,t}$ for target point cloud $\mathbf{Q}$.

\subsection{Self-Supervised Loss Functions}\label{loss}
At each iteration, we can obtain the estimated flow $\mathbf{F}_t = \{\mathbf{F}^{p,t},\mathbf{F}^{q,t}\}$ for pairwise point clouds $\mathbf{P}$ and $\mathbf{Q}$.
Without the ground truth scene flow, we utilize the following loss functions for model training. For simplicity, we omit the iteration subscript.
 
\noindent \textbf{Chamfer loss.}
Following \cite{wu2020pointpwc,kittenplon2021flowstep3d}, we warp the source $\mathbf{P}$ with the predicted flow $\mathbf{F}^p$ and minimize the Chamfer Distance between the warped source $\bm{P}'$ and target $\mathbf{Q}$ by
\begin{align}\label{eq19}
L_{ch}(\mathbf{P'},\mathbf{Q}) = &\sum_{\mathbf{q}_j \in \bm{Q}}\mathop{\min}_{\mathbf{p'}_i \in    \mathbf{P'}}\|\mathbf{q}_j - \mathbf{p'}_i \|_{2} + \nonumber\\
& \sum_{\mathbf{p'}_i \in \mathbf{P'}}\mathop{\min}_{\mathbf{q}_j \in \mathbf{Q}}\|\mathbf{p'}_i - \mathbf{q}_j \|_{2}
\end{align}

\noindent \textbf{Smoothness loss.}
Following \cite{wu2020pointpwc,kittenplon2021flowstep3d}, we also constrain the predicted scene flow values within a  small local region to be similar. The smoothness loss is defined as
  \begin{equation}
L_{s} = \sum_{\mathbf{p}_i \in \bm{P}}\frac{1}{|\mathcal{N}'_i|}\sum_{\mathbf{p}_j \in \mathcal{N}'_i}\|\mathbf{f}^p_i - \mathbf{f}^p_j \|_{2}
\end{equation}
where $\mathcal{N}'_i$ is the neighborhood around $\mathbf{p}_i \in \bm{P}$.

\noindent \textbf{Consistency loss.}
We enforce the backward-interpolated flow of the target point clouds to be consistent with the predicted flow of the source point clouds and vice versa.
\begin{equation}
L_{c} =  \|\mathbf{F}^p - \Omega\left(\mathbf{F}^q \right)\|_2 + \|\mathbf{F}^q - \Omega\left(\mathbf{F}^p \right)\|_2
\end{equation}
where $\Omega$ is backward interpolation.

The combined loss for self-supervised training can be written as
\begin{align}
& L = L_{ch}  + \alpha L_{s}  + \beta L_{c}
\end{align}
where $\alpha$ and $\beta$ are the regularization parameters.

\section{Experiment}

\subsection{Experimental Setups}\label{setups}
\noindent \textbf{Datasets.} 
To validate the effectiveness of our proposed scene flow estimation framework, we conduct extensive experiments on two benchmarks, the FlyingThings3D \cite{mayer2016large} and the KITTI Scene Flow \cite{menze2015joint,menze2018object}. 
There are two versions of datasets.
The first version of the datasets is prepared 
by HPLFlowNet \cite{gu2019hplflownet}. We denote these datasets without occluded points $\operatorname{FT3D_s}$ and $\operatorname{KITTI_s}$, respectively.
$\operatorname{FT3D_s}$ contains 19640 training examples and 3824 pairs in the test set. We only use one-quarter of the training data (4910 pairs). $\operatorname{KITTI_s}$ is a real-world scene flow dataset with 200 pairs for which 142 are used for testing without any fine-tuning. 
The second version of the datasets is prepared 
by FlowNet3D \cite{liu2019flownet3d}. This version of datasets includes the occluded points, which are denoted $\operatorname{FT3D_o}$ and $\operatorname{KITTI_o}$, respectively. 
$\operatorname{FT3D_o}$ contains 19999 training examples and 2003 pairs in the test set. $\operatorname{KITTI_o}$ consists of 150 test examples. 
Besides, following Self-Point-Flow \cite{li2021self}, we also split the $\operatorname{KITTI_o}$ dataset into $\operatorname{KITTI_f}$ with 100 pairs and $\operatorname{KITTI_t}$ with 50 pairs for evaluation. Moreover, \cite{li2021self} also extracts another self-supervised training dataset with 6026 pairs from the original KITTI dataset, denoted as $\operatorname{KITTI_r}$. 

\noindent \textbf{Implementation details.} 
Our model is implemented with Pytorch and all experiments are executed on a NVIDIA TITAN RTX GPU. 
For the experiments on point clouds without occlusions, we train our model on synthetic $\operatorname{FT3D_s}$ training data and evaluate it on both $\operatorname{FT3D_s}$ test data and $\operatorname{KITTI_s}$ dataset. we feed randomly sampled 8192 points as inputs to our model, just like \cite{puy2020flot,li2022rigidflow} and other compared methods, and train it with a batch size of 2. Besides, we set the superpoint number and iteration number to 128 and 3, respectively.
For occluded experiments, like \cite{li2022rigidflow}, we also train our model on $\operatorname{KITTI_r}$ dataset and test it on $\operatorname{KITTI_o}$ and $\operatorname{KITTI_t}$. The size of the input point clouds is set to 2048. Here, we set the batch size, iteration number, and superpoint number to 4, 3, and 30, respectively.
The initial learning rate used in all experiments is 0.001 and our model is optimized with the ADAM optimizer. We multiply the learning rate by 0.7 at epochs 40, 55, and 70 and train our model for 100 epochs.

\noindent \textbf{Evaluation metrics.} 
We test our model with four evaluation metrics used in \cite{liu2019flownet3d,gu2019hplflownet}, including End Point Error (EPE), Accuracy Strict (AS), Accuracy Relax (AR), and Outliers (Out). We denote the estimated scene flow and ground truth scene flow as $\mathbf{F}$ and $\mathbf{F}_{gt}$, respectively. EPE(m): $||\mathbf{F} - \mathbf{F}_{gt} ||_2$ averaged over all points.
AS(\%): the percentage of points whose EPE $\textless 0.05m$ or relative error $\textless 5\%$. AR(\%): the percentage of points whose EPE $\textless 0.1m$ or relative error $\textless 10\%$. Out(\%): the percentage of points whose EPE $\textgreater 0.3m$ or relative error $\textgreater 10\%$.

\subsection{Results}\label{occluded}
\noindent \textbf{Performance on point clouds without occlusions.}
We train our self-supervised model on  $\operatorname{FT3D_s}$ training data and evaluate it on both $\operatorname{FT3D_s}$ test data and $\operatorname{KITTI_s}$ dataset. And we compare our model with the recent state-of-the-art self-supervised scene flow estimation methods, including Ego-Motion \cite{tishchenko2020self}, PointPWC-Net \cite{wu2020pointpwc}, SLIM \cite{baur2021slim}, Self-Point-Flow \cite{li2021self}, FlowStep3D \cite{kittenplon2021flowstep3d}, RCP \cite{gu2022rcp}, PDF-Flow \cite{he2022self}, and RigidFlow \cite{li2022rigidflow}. The results are reported in Table \ref{table0}. From the results, it can be found that our model can outperform all compared self-supervised methods in terms of the four metrics on the $\operatorname{FT3D_s}$ test data. Especially, our model brings 8.72\% gains for metric AS. For the $\operatorname{KITTI_s}$ dataset, our model brings substantial improvements on all metrics. To be specific, our model outperforms the second best method RCP \cite{gu2022rcp} by 8.68\% and 6.58\% on metrics AS and AR, respectively. Besides, it is worth noting that our model can even achieve an EPE metric of 3.62$cm$, which is much lower than the EPE (6.19$cm$) of recent RigidFlow. 

We also compare our model with some supervised methods, such as FlowNet3D \cite{liu2019flownet3d} and FLOT \cite{puy2020flot}, etc. As shown in Table \ref{table0}, our self-supervised model achieves comparable performance with supervised HPLFlowNet \cite{gu2019hplflownet} on $\operatorname{FT3D_s}$ dataset. 
Without any fine-tuning on $\operatorname{KITTI_s}$ dataset, our model can even outperform the supervised methods listed in Table \ref{table0}, which proves that our model has better generalization ability. For real scenes, most local regions are with similar flow patterns.  
Thanks to dynamically clustering mechanism, our model clusters points with similar flow pattern into the same clusters and encodes the superpoint-level flow into the GRU for flow refinement, thereby leading to satisfactory performance on real scenes. 

\begin{table}[t]
	\centering
	\setlength{\tabcolsep}{1.2mm}{
		\begin{tabular}{l|c|cccc}
			\toprule
			Methods&Sup.&EPE $\downarrow$ & AS $\uparrow$ & AR $\uparrow$  &  Out $\downarrow$ \\
			\midrule \midrule
			\multicolumn{6}{c}{$\operatorname{FT3D_s}$} \\
			\midrule
			FlowNet3D \small\cite{liu2019flownet3d}&Full.&  0.1136 &  41.25  &  77.06  & 60.16   \\
			HPLFlowNet \small\cite{gu2019hplflownet}&Full.   & 0.0804  & 61.44   & 85.55   & 42.87   \\
			PointPWC-Net \small\cite{wu2020pointpwc}&Full. & 0.0588  &  73.79  &  92.76  &  34.24 \\
			\cline{1-6}	
			Ego-Motion \small\cite{tishchenko2020self}&Self. & 0.1696  & 25.32  &  55.01  & 80.46  \\	
			PoinPWC-Net \small\cite{wu2020pointpwc}&Self. &  0.1246 &  30.68  &  65.52  & 70.32  \\	
			Self-Point-Flow \small\cite{li2021self}&Self. & 0.1009  & 42.31   &  77.47  &  60.58 \\
			FlowStep3D \small\cite{kittenplon2021flowstep3d}&Self. & 0.0852  & 53.63   &  82.62  &  41.98\\
			PDF-Flow \small\cite{he2022self}&Self. & 0.075  & 58.9  & 86.2  & 47.0  \\	
			RCP \small\cite{gu2022rcp}&Self. & 0.0765  &  58.58  &   86.02 & \underline{41.42}  \\
			RigidFlow \small\cite{li2022rigidflow}&Self. &  \underline{0.0692} &  \underline{59.62}  &  \underline{87.10}  &  46.42 \\
			SPFlowNet (ours)&Self. & \textbf{0.0606}  &  \textbf{68.34}  &   \textbf{90.74} & \textbf{38.76} \\
			\midrule
			\multicolumn{6}{c}{$\operatorname{KITTI_s}$}             \\
			\midrule
			FlowNet3D \small\cite{liu2019flownet3d}&Full.& 0.1767  & 37.38 & 66.77  & 52.71 \\
			HPLFlowNet \small\cite{gu2019hplflownet}&Full.   & 0.1169  & 47.83  & 77.76  & 41.03   \\
			PointPWC-Net \small\cite{wu2020pointpwc}&Full.& 0.0694  & 72.81  & 88.84  &  26.48  \\
			FLOT \small\cite{puy2020flot}&Full.& 0.0560  & 75.50  &  90.80 & 24.20   \\
			\cline{1-6}
			Ego-Motion \small\cite{tishchenko2020self}&Self. &  0.4154 &  22.09  &  37.21  & 80.96  \\	
			PoinPWC-Net \small\cite{wu2020pointpwc}&Self. &  0.2549 &  23.79  &  49.57  &  68.63 \\	
			SLIM \small\cite{baur2021slim}&Self. &  0.1207 &  51.78  &  79.56  &  40.24 \\	
			FlowStep3D \small\cite{kittenplon2021flowstep3d}&Self. & 0.1021  & 70.80   & 83.94   & 24.56  \\	
			PDF-Flow \small\cite{he2022self}&Self. & 0.092  & 74.7  & 87.0  & 28.3 \\
			Self-Point-Flow \small\cite{li2021self}&Self. & 0.1120  &   52.76 &  79.36  & 40.86  \\
			RigidFlow \small\cite{li2022rigidflow}&Self. & \underline{0.0619}  &  72.37  & 89.23   & 26.18  \\
			RCP \small\cite{gu2022rcp}&Self. &  0.0763 &  \underline{78.56}  & \underline{89.21}   &  \underline{18.49} \\
			SPFlowNet (ours)&Self. &\textbf{0.0362}  &  \textbf{87.24}  &  \textbf{95.79} &  \textbf{17.71} \\
			\bottomrule
		\end{tabular}
		\caption{Comparison results on the $\operatorname{FT3D_s}$ and $\operatorname{KITTI_s}$ datasets. Our model is trained on $\operatorname{FT3D_s}$ training part and evaluated on $\operatorname{FT3D_s}$ test set and $\operatorname{KITTI_s}$ dataset. Full. means the fully-supervised training manner. Self. represents the self-supervised training manner. Note that the best and the second-best results are emboldened and underlined, respectively.}
		\label{table0}}
\end{table}

\begin{table}[ht]
	\centering
	\setlength{\tabcolsep}{0.5mm}{
		\begin{tabular}{l|l|c|cccc}
			\toprule
			Methods&Sup. & T. data&EPE $\downarrow$ & AS $\uparrow$ & AR $\uparrow$  &  Out $\downarrow$ \\
			\midrule 
			FlowNet3D \small\cite{liu2019flownet3d}&Full.&  $\rm{F_o}$ & 0.173 &  27.6 & 60.9 & 64.9   \\
			FLOT \small\cite{puy2020flot}&Full.   &  $\rm{F_o}$ & 0.107 &  45.1 & 74.0 & 46.3  \\
			Self-Point-Flow \small\cite{li2021self}&Self. & $\rm{K_r}$  & 0.105 & 41.7  & 72.5 &  50.1 \\
			RigidFlow \small\cite{li2022rigidflow}&Self. &  $\rm{K_r}$ & \underline{0.102} &  \underline{48.4} & \underline{75.6} &\underline{44.2} \\
			SPFlowNet (ours)&Self. &$\rm{K_r}$  &  \textbf{0.086}  &  \textbf{61.1} &  \textbf{82.4} &  \textbf{39.1} \\
			\bottomrule
		\end{tabular}
		\caption{Comparison results on $\operatorname{KITTI_o}$ dataset. Our model is trained on $\operatorname{KITTI_r}$ and evaluated on $\operatorname{KITTI_o}$ dataset. T. data: training data. $\operatorname{F_o}$: $\operatorname{FT3D_o}$. $\operatorname{K_r}$:$\operatorname{KITTI_r}$.}
		\label{table1}}
\end{table}

\noindent \textbf{Performance on point clouds with occlusions.}
Following the experimental settings used in Self-Point-Flow \cite{li2021self} and RigidFlow \cite{li2022rigidflow}, we train our model on $\operatorname{KITTI_r}$ dataset and evaluate our model on both $\operatorname{KITTI_o}$ and $\operatorname{KITTI_t}$ datasets. The results on $\operatorname{KITTI_o}$ and $\operatorname{KITTI_t}$ are shown in Tables \ref{table1} and \ref{table2}, respectively. Although our model is not designed to deal with occluded cases, our model can also achieve the best performance on $\operatorname{KITTI_o}$ dataset.
This is due to that although there is no correspondence of the occluded points, our model employs the superpoint-level flow to guide the flow refinement rather than point-level flow information, which can alleviate the occluded problem to some extent.
Due to the lack of point-level flow annotations for the real scenes, the supervised FLOT and FlowNet3D are trained on synthetic $\operatorname{FT3D_o}$ dataset. The other two self-supervised methods \cite{li2021self, li2022rigidflow} and our model can be trained directly on unlabeled outdoor $\operatorname{KITTI_r}$ dataset.
As shown in Table \ref{table1}, our model can outperform all self-supervised methods including Self-Point-Flow and RigidFlow.
To be specific, our model brings 12.7\% gains on metric AS. Besides, it is worth noting that the Self-Point-Flow \cite{li2021self} needs additional normal and color information for pseudo label generation. Our model only needs the coordinate information of the consecutive frames of point clouds as inputs.
For the $\operatorname{KITTI_t}$ dataset, we compare our model with JGF \cite{mittal2020just} and WWL \cite{pontes2020scene}. These two methods use the pre-trained model of FlowNet3D on $\operatorname{FT3D_o}$ as the baseline and perform self-supervised fine-tuning on $\operatorname{KITTI_f}$, and then test their model on $\operatorname{KITTI_t}$ dataset. Our model and RigidFlow \cite{li2022rigidflow} get rid of the pre-trained model on synthetic $\operatorname{FT3D_o}$ and only need to be trained on the unlabeled $\operatorname{KITTI_r}$ in a self-supervised manner. 
As shown in Table \ref{table2}, our model can obtain 9.92\% improvements on metric AR. 

\begin{table}[ht]
	\centering
	\setlength{\tabcolsep}{1.0mm}{
		\begin{tabular}{l|c|c|ccc}
			\toprule
			Methods&Pre-T.& T. data&EPE $\downarrow$ & AS $\uparrow$ & AR $\uparrow$ \\
			\midrule 
			TGF \small\cite{mittal2020just}& $\checkmark$ &  $\rm{F_o} + \rm{K_f}$ & 0.218 &  10.17 & 34.38 \\
			WWL \small\cite{pontes2020scene}&  $\checkmark$ &  $\rm{F_o} + \rm{K_f}$ & 0.169 & 21.71  & 47.75\\
			RigidFlow \small\cite{li2022rigidflow}& &  $\rm{K_r}$ & \underline{0.117} & \underline{38.75} & \underline{69.73}\\
			SPFlowNet (ours)& &$\rm{K_r}$  &  \textbf{0.089}  &  \textbf{53.28} &  \textbf{79.65} \\
			\bottomrule
		\end{tabular}
		\caption{Comparison results on $\operatorname{KITTI_t}$ dataset. Our model is trained on $\operatorname{KITTI_r}$ and evaluated on $\operatorname{KITTI_t}$ dataset. 
	}
		\label{table2}}
\end{table}
\subsection{Ablation Study}\label{ablation}
\noindent \textbf{The effectiveness of key components.} We conduct experiments to verify the effectiveness of key components in our model. 
Firstly, we remove the superpoint generation and superpoint guided flow refinement modules in our model. This variant takes a GRU without superpoint guidance for flow refinement (abbr. as ``w/o superpoint''). Secondly, we adopt the SPNet \cite{hui2021superpoint} for superpoint generation without flow guidance (abbr. as ``w/ SPNet''). The model ``w/ FGSG (ours)'' represents our model with flow guided superpoint generation module. The results of the above three models are listed in the top part of Table \ref{table3}. From the results of the variant ``w/o superpoint'' and the other two models with superpoints, it can be found that introducing superpoints into scene flow estimation is effective. Besides, our proposed flow guided superpoint generation module can achieve better results than SPNet, which shows that flow guidance is crucial when there is no ground truth superpoint labels. 
Besides, we remove the consistency encoding from our model (abbr. as ``w/o cons. encoding'').
Table \ref{table3} shows that the performance drops a lot without the superpoint consistency encoding, which demonstrates that the consistency between the reconstructed superpoint-level flow of pairwise point clouds is important.
Finally, we also remove the consistency loss and only utilize the Chamfer loss and smoothness loss for model training (abbr. as ``w/o cons. loss''). The results of our model without consistency loss are worse than with it. According to the above comparisons, it can be observed that our model is less effective without any key components.

\begin{table}[t]
	\centering
	\setlength{\tabcolsep}{2.4mm}{
		\begin{tabular}{l|cccc}
			\toprule
			Methods&EPE $\downarrow$ & AS $\uparrow$ & AR $\uparrow$  &  Out $\downarrow$ \\
			\midrule
			w/o superpoint&  0.119 &   55.4 & 72.9 & 45.2\\
			w/ SPNet &  0.090&   60.0 & 80.7 & 40.2\\
			w/ FGSG (ours)&\textbf{0.086}  &  \textbf{61.1}  &  \textbf{82.4} &  \textbf{39.1} \\
			\midrule
			w/o cons. encoding &  0.103&   57.7 & 76.1 & 44.3\\	
			w/o cons. loss& 0.094 &  59.0  & 80.0 & 40.7\\
			SPFlowNet (ours)&\textbf{0.086}  &  \textbf{61.1}  &  \textbf{82.4} &  \textbf{39.1} \\
			\bottomrule
		\end{tabular}
		\caption{Comparison results on the $\operatorname{KITTI_o}$ dataset. All models are trained on $\operatorname{KITTI_r}$ and evaluated on $\operatorname{KITTI_o}$ dataset.}
		\label{table3}}
\end{table}

\begin{figure}[t]
	\centering
	\includegraphics[width=0.45\textwidth]{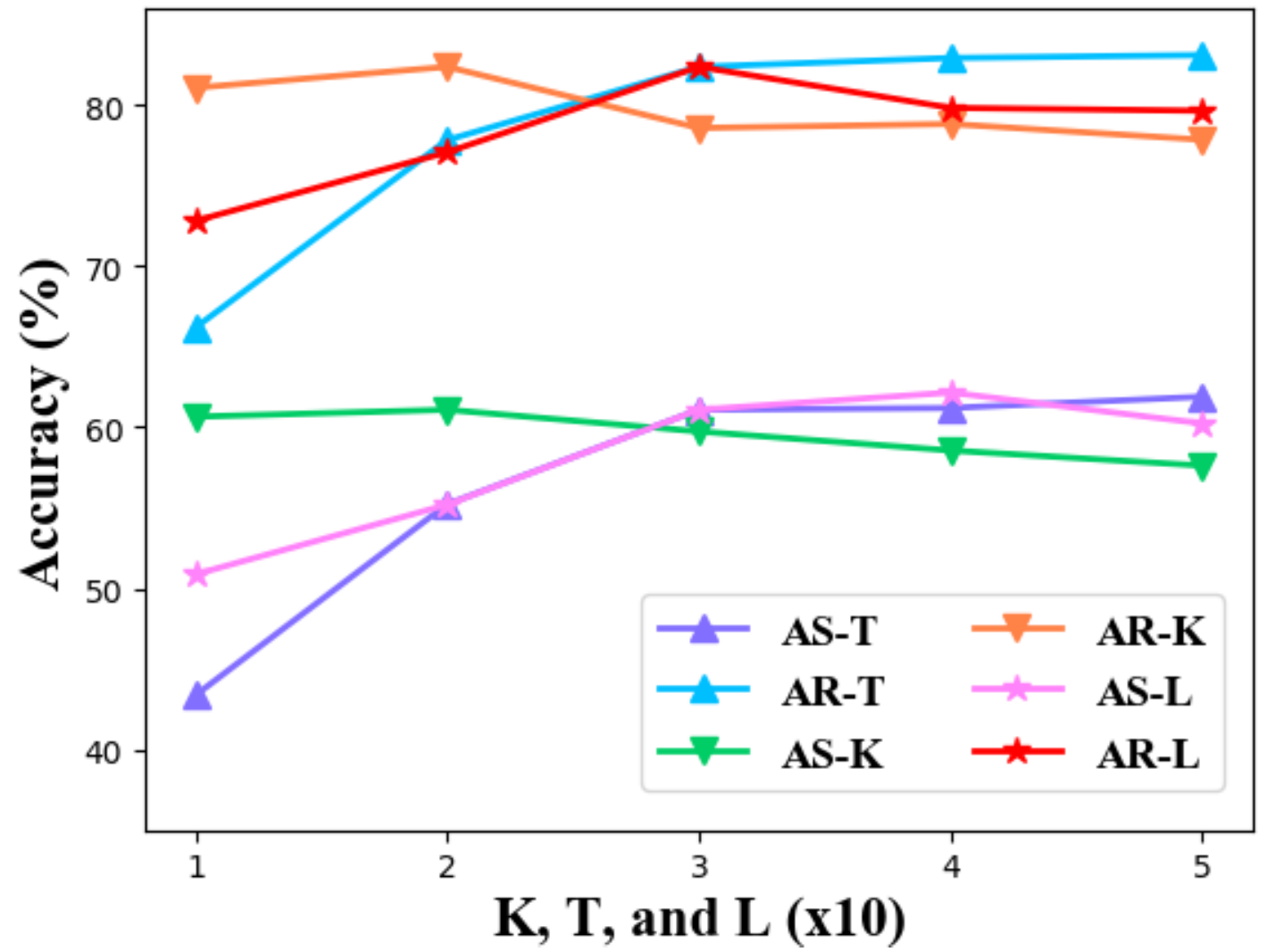}
	\vspace{-5pt}
	\caption{The ablation study results (\textbf{AS} and \textbf{AR}) of different hyper-parameters $L$, $K$, and $T$ on the $\operatorname{KITTI_o}$ dataset, where $L \in \{10,20,30,40,50\}$ and $K,T \in \{1,2,3,4,5\}$.}
	\label{image2}
\end{figure}

\noindent \textbf{Choices of the superpoint number $L$.} In our superpoint generation layer, we generate $L$ superpoints. We conduct the ablation study to choose a suitable superpoint number. We plot the results of the metrics AS and AR with different $L \in \{10,20,30,40,50\}$ in Figure \ref{image2}. It can be observed that choosing $L=30$ achieves the best results.  

\noindent \textbf{Impact of the $K$-nearest superpoint centers.} To prevent a point from being clustered to a distant superpoint, we only calculate the association map between each point and its $K$-nearest superpoint centers. Here we explore the impact on the performance of different $K$. We fix other super-parameters and choice $K \in \{1,2,3,4,5\}$. The accuracy results are visualized in Figure \ref{image2}. Figure \ref{image2} shows that our model achieves the best performance with $K=2$.

\noindent \textbf{Number of iterations $T$.} Our model iteratively generates superpoints and conducts the superpoint guided flow refinement. We plot the accuracy results of our model after each iteration. From Figure \ref{image2}, it can be found that $T=3$ can obtain state-of-the-art performance. Although $T=4,5$ can achieve slightly high accuracy, it increases the inference time. Therefore, for a good trade-off between the accuracy and efficiency, we choose $T=3$.

\section{Conclusion}
We proposed a novel end-to-end superpoint guided scene flow estimation framework. Different from other offline clustering based scene flow estimation methods, our method can simultaneously optimize the flow guided superpoint generation and superpoint guided flow refinement. Thanks to the joint end-to-end optimization, our model can gradually generate more accurate flow results.
Extensive experiments on the synthetic and real LiDAR scenes demonstrate that our self-supervised model can achieve outstanding performance in the scene flow estimation task.

{\small
\bibliographystyle{ieee_fullname}
\bibliography{egbib}
}

\end{document}